\documentclass[sigconf]{acmart}
\AtBeginDocument{%
  }

\setcopyright{acmlicensed}

\copyrightyear{2026}
\acmYear{2026}
\setcopyright{cc}
\setcctype{by}
\acmConference[WWW '26] {Proceedings of the ACM Web Conference 2026}{April 13--17, 2026}{Dubai, United Arab Emirates.}
\acmBooktitle{Proceedings of the ACM Web Conference 2026 (WWW '26), April 13--17, 2026, Dubai, United Arab Emirates}
\acmISBN{979-8-4007-2307-0/2026/04}
\acmDOI{10.1145/3774904.3792245}

\acmSubmissionID{rfp1160}

\usepackage{xspace}
\usepackage{lipsum}
\usepackage{booktabs}
\usepackage{tabularx}
\usepackage{caption}
\usepackage{multirow}
\usepackage{graphicx}
\usepackage{colortbl}
\usepackage{xcolor} 
\usepackage{array}
\usepackage[table]{xcolor}

\begin{document}

\title{Personalized Learning Path Planning through Goal-Driven\\Learner State Modeling}


\author{Joy Jia Yin Lim}
\orcid{0009-0006-7971-6096}
\email{lin-jy23@mails.tsinghua.edu.cn}
\affiliation{
  \department{Department of Computer Science and Technology, Beijing National Research Center for Information Science and Technology}
  \institution{Tsinghua University}
  \city{Beijing}
  \country{China}
}

\author{Ye He}
\orcid{0009-0001-8772-1104}
\email{heye23@mails.tsinghua.edu.cn}
\affiliation{
  \department{Department of Computer Science and Technology}
  \institution{Tsinghua University}
  \city{Beijing}
  \country{China}
}

\author{Jifan Yu}
\orcid{0000-0003-3430-4048}
\email{yujifan@tsinghua.edu.cn}
\affiliation{
  \department{Institution of Education}
  \institution{Tsinghua University}
  \city{Beijing}
  \country{China}
}

\author{Xin Cong}
\orcid{0000-0003-2370-306X}
\email{congxin1995@tsinghua.edu.cn}
\affiliation{
  \department{Department of Statistics and Data Science}
  \institution{Tsinghua University}
  \city{Beijing}
  \country{China}
}

\author{Daniel Zhang-Li}
\orcid{0009-0009-3681-1896}
\email{zlnn23@mails.tsinghua.edu.cn}
\affiliation{
  \department{Department of Computer Science and Technology}
  \institution{Tsinghua University}
  \city{Beijing}
  \country{China}
}

\author{Zhiyuan Liu}
\orcid{0000-0002-7709-2543}
\email{liuzy@tsinghua.edu.cn}
\affiliation{
  \department{Department of Computer Science and Technology}
  \institution{Tsinghua University}
  \city{Beijing}
  \country{China}
}

\author{Huiqin Liu}
\orcid{0000-0002-5754-2623}
\email{liuhq@tsinghua.edu.cn}
\affiliation{
  \department{Institution of Education}
  \institution{Tsinghua University}
  \city{Beijing}
  \country{China}
}

\author{Lei Hou}
\orcid{0000-0002-8907-3526}
\email{houlei@tsinghua.edu.cn}
\author{Juanzi Li}
\orcid{0000-0002-6244-0664}
\email{lijuanzi@tsinghua.edu.cn}
\affiliation{
  \department{Department of Computer Science and Technology}
  \institution{Tsinghua University}
  \city{Beijing}
  \country{China}
}

\author{Bin Xu}
\orcid{0000-0003-3040-4391}
\authornotemark[2]
\email{xubin@tsinghua.edu.cn}
\affiliation{
  \department{Department of Computer Science and Technology, Beijing National Research Center for Information Science and Technology}
  \institution{Tsinghua University}
  \city{Beijing}
  \country{China}
}

\renewcommand{\shortauthors}{Joy Jia Yin Lim et al.}
\renewcommand{\shorttitle}{Personalized Learning Path Planning through Goal-Driven Learner State Modeling}
\newcommand{\system}{{\textit{Pxplore}}\xspace}

\begin{abstract}
Personalized Learning Path Planning (PLPP) aims to design adaptive learning paths that align with individual goals. While large language models (LLMs) show potential in personalizing learning experiences, existing approaches often lack mechanisms for goal-aligned planning. We introduce \system, a novel framework for PLPP that integrates a reinforcement-based training paradigm and an LLM-driven educational architecture. We design a structured learner state model and an automated reward function that transforms abstract objectives into computable signals. We train the policy combining supervised fine-tuning (SFT) and Group Relative Policy Optimization (GRPO), and deploy it within a real-world learning platform. Extensive experiments validate \system's effectiveness in producing coherent, personalized, and goal-driven learning paths. We release our code and dataset at ~\url{https://github.com/Pxplore/pxplore-algo}.

\end{abstract}



\maketitle

\section{Introduction}

\begin{figure}[t]
  \centering
  \includegraphics[width=0.9\columnwidth]{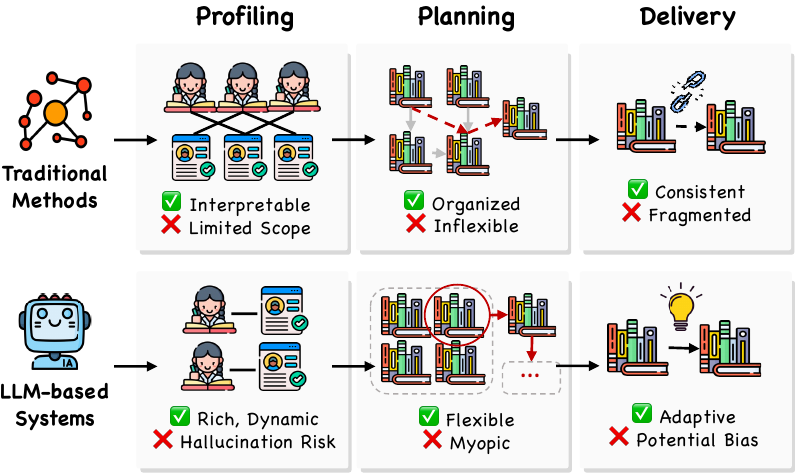}
  \caption{Traditional methods produce static paths, while LLMs struggle to capture long-term progress.}
  \label{fig:intro}
\end{figure}

Learning is a sequential and cumulative process that varies across learner. \textbf{Personalized learning path planning (PLPP)} for each learner has been a central pursuit in education ~\cite{ng2024educational,maghsudi2021personalized}. As shown in Figure~\ref{fig:intro}, traditional approaches to PLPP~\cite{tiwari2023integration,de2023exploring}, such as collaborative filtering and knowledge graphs, have advanced the organization of heterogeneous learning resources. However, their personalization remains constrained by predefined resources and static learner profiles, leading to fragmented and learning experiences~\cite{ji2024genrec,abu2024knowledge,li2023research}. Large language models (LLMs), in contrast, overcome these limitations through scalable learner modeling and dynamic content adaptation~\cite{wen2024ai,hu2024dynamic,hu2024teaching}. They integrates multi-dimensional learner data, adapts instructional materials, and even generates new content in a fine-grained, context-aware manner~\cite{Wang_2025,chen2024large,fitrianto2024utilizing}. Yet, despite this potential, LLMs still struggle to model sustained learning progression, as they often lack mechanisms to represent long-term learner development or to incorporate feedback from authentic educational interactions~\cite{jung2024personalized,zhao2024learning}.

These limitations highlight two fundamental challenges.

$\bullet$ \textbf{Planning Long-term  Learning Process}. Learning is a goal-driven, multi-step process. Existing LLM-based systems are often driven by short-sighted or single-step decisions~\cite{cevikbas2024empirical,duha2023chatgpt,whalen2023chatgpt}, without optimizing for cumulative educational impact. Effective long-term learning path planning requires a general mechanism to capture the abstract educational goals and translate them into computable reward signals that can guide further optimization~\cite{fahad2023reinforcement}.

$\bullet$ \textbf{Understanding diverse learner states and adapting to heterogeneous learning resources}. Learner states, evolving continuously throughout the learning process~\cite{an2025modeling}, requires adaptive systems that can infer and respond to these dynamic changes. However, existing approaches often lack explicit mechanisms to model such evolving learner states or to integrate diverse learning resources~\cite{hu2024teaching,moundridou2024generative}, limiting their ability to construct and deliver coherent personalized learning experiences~\cite{ikwuanusi2023ai,bernacki2021systematic,maghsudi2021personalized}.

In this paper, we introduce \system, a structured framework for PLPP that integrates a reinforcement-based training paradigm and an LLM-driven educational architecture. We first define a goal-driven learner state model to represent learning objectives and learning motivations. Building on this, we design an automated reward function that transforms these abstract states into computable reward signals by quantifying the alignment of objectives and motivations in a learning process. This signal guides a two-stage training pipeline, including supervised fine-tuning (SFT) for initialization and Group Relative Policy Optimization (GRPO)~\cite{shao2024grpo} for further refinement. We then deploy the policy into an LLM-driven educational architecture, unifying learner profiling, path planning, and adaptive delivery into an integrated learning process.

Our contributions can be summarized as follows:

(1) We propose a training paradigm that integrates goal-driven learner state modeling with reinforcement-based optimization.

(2) We design an integrated system architecture that enables real-world deployment with coherent profiling, planning, and delivery.

(3) We conduct extensive experiments and user studies to validate the practical utility of our proposed system.

\begin{figure*}[t]
  \centering
\includegraphics[width=0.9\linewidth]{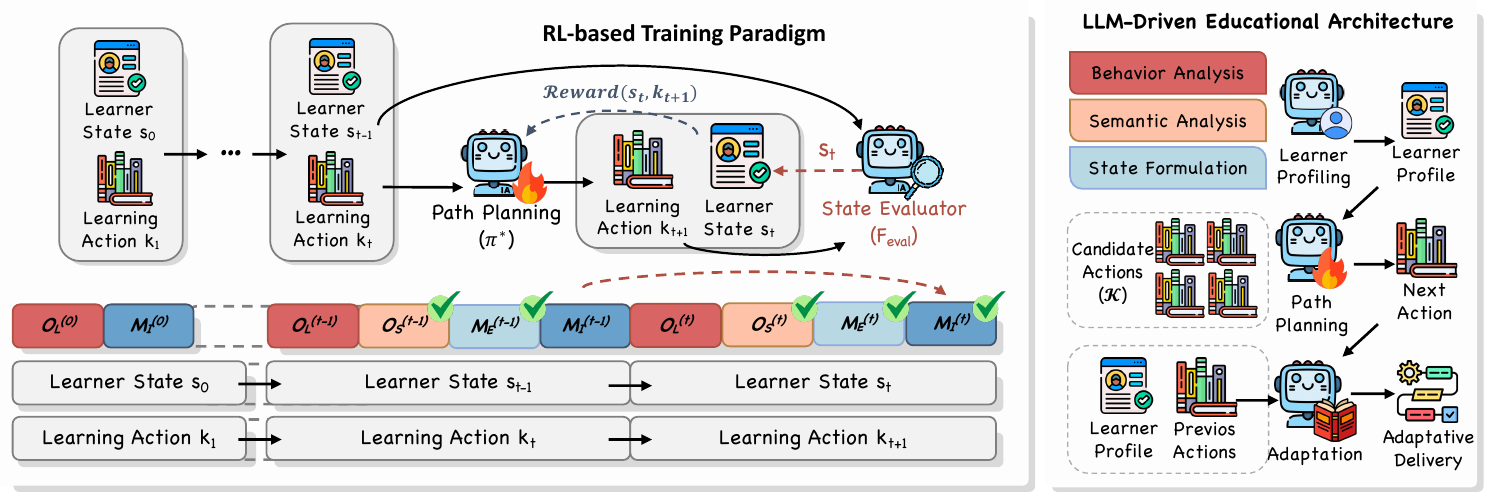}
  \caption{Overview of the \system framework, including: (1) an Reinforcement-based Training Paradigm for goal-aligned learning path planning; and (2) an LLM-driven educational architecture for adaptive profiling, planning and delivery.}
  \label{fig:pxplore_framework}
\end{figure*}

\section{Related Work}

\textbf{$\bullet$ Personalized Learning Path Planning.}
Personalized Learning Path Planning (PLPP) has long been a central objective in education~\cite{ng2024educational}. Early research focused on adapting learning resources through data-driven approaches such as collaborative filtering, which recommends materials by identifying learners with similar profiles and learning behaviors~\cite{harpale2008personalized,cui2020personalized,elshani2021constructingpersonalizedlearningpath}. Another line of work leveraged knowledge graph to model relationships among concepts, allowing systems to infer prerequisite structures and suggest coherent learning sequence within a curriculum~\cite{abu2023knowledgegraphs,chen2018knowedu}. 
While these methods laid the groundwork for personalization, they were largely constrained to single-step adaptations and surface-level content alignment. Recently, advances in LLMs has revitalized this area~\cite{xiao2025personalized,chen2025research}. Researchers have begun to generate personalized and context-aware learning experiences at scale~\cite{zhao2025new,chu2025llm,yusuf2025pedagogical,dai2024effects}, moving beyond static resource organization toward more adaptive, goal-driven, and instructional coherent learning support.  

\textbf{$\bullet$ Reinforcement Learning for Education}
Reinforcement learning (RL) provides a principled framework for goal-driven planning and sequential decision-making \cite{memarian2024scoping,singla2021reinforcementlearningeducationopportunities}, making it well-suited for PLPP. As learning involves sequences of learning actions rather than isolated decisions, RL has been explored to optimize instructional policies~\cite{Zhao_2024,10499810,YUN2024111242} and simulate learner behaviors~\cite{2025Scarlatos,yu2024systematic} in dynamic environments. 
Despite these advances, applying standard RL to real-world settings remains challenging~\cite{memarian2024scoping}. Learning paths are heterogeneous, long-horizon, and driven by abstract objectives, such as curiosity and deep understanding, which extend beyond measurable task performance~\cite{singla2021reinforcementlearningeducationopportunities}. This highlight the need for novel RL paradigms that can represent abstract, evolving objectives and operate effectively across diverse, real-world contexts.

\section{\system Framework}

The \system framework unifies an \textbf{reinforcement-based training paradigm} with an \textbf{LLM-based educational architecture}. 

\subsection{Reinforcement-based Training Paradigm}
While LLMs demonstrate strong planning and reasoning capabilities under RL frameworks such as GRPO, their application to open-ended domains remains challenging. The inherent complexity of multi-dimensional objectives, sparse feedback, and evolving learner goals requires specialized adaptations. To address this, we design an educational adaptation of GRPO that integrates structured learner state representation and abstract reward modeling, enabling robust long-horizon optimization in learning path planning.

\subsubsection{Problem Formulation}

We formalize PLPP as a sequential decision-making process governed by complex state and abstract rewards. We define the foundational elements as follows:

\emph{\textbf{Learner State ($s_t$).}}
A learner state $s_t \in \mathcal{S}$ is a structured representation inferred from learner interactions at time $t$, where $\mathcal{S}$ denotes the state space. This state captures both explicit cognitive attributes (e.g., knowledge mastery, misconceptions) and implicit factors (e.g., motivation, goals, and engagement). 

\emph{\textbf{Learning Action ($k$).}}
Given a knowledge corpus $\mathcal{K}$, an action $k_t \in \mathcal{K}$ is an atomic unit of instruction selected at timestep $t$, such as a concept explanation, worked example, or short exercise. 

\emph{\textbf{Learning Path ($P_T$).}}
A learning path $P_T$ is a finite, time-ordered sequence of learning actions over $T$ steps, $P_T = (k_1,\dots, k_T)$, which represents the learning path guided by evolving learner states.

\emph{\textbf{Personalized Learning Path Planning (PLPP).}}
The objective of PLPP is to learn an optimal policy, $\pi^*$, that maps a learner's current state $s_t$ to a distribution over possible next actions $k_{t+1}$. Formally, a policy is a mapping
\begin{equation}
\pi: \mathcal{S} \to \Delta(\mathcal{K}),
\end{equation}
where $\Delta(\mathcal{K})$ denotes the probability simplex over the corpus actions $\mathcal{K}$. At each step $t$, the system samples an action and updates the state according to a transition dynamic $T$:
\begin{equation}
k_{t+1} \sim \pi(\cdot | s_t), \qquad s_{t+1} \sim T(\cdot | s_t, k_{t+1}).
\end{equation}
The goal is to maximize expected cumulative pedagogical reward
\begin{equation}
\pi^* = \arg\max_{\pi} \mathbb{E}{\pi}!\left[\sum{t=1}^{T} \gamma^{t-1} \mathcal{R}(s_t, k_{t+1})\right],
\end{equation}
where $\mathcal{R}$ is scalar reward and $\gamma \in [0,1]$ is temporal discount factor.

\subsubsection{Abstract State Modeling}
\label{sec:learner_state_model}

To enable policy optimization over abstract pedagogical goals, we introduce a structured state representation, the \textbf{Learner State Model}. Grounded in Goal-Setting Theory~\cite{Locke1990} and Achievement Goal Theory~\cite{chi2004achievement}, it models both cognitive and motivational aspects of learner behavior. At timestep $t$, a learner state is represented as
\begin{equation}
\label{eq:state_structure}
s_t = (O_L^{(t)}, O_S^{(t)}, M_I^{(t)}, M_E^{(t)}),
\end{equation}
where $O_L^{(t)}$ and $O_S^{(t)}$ denote the sets of \emph{long-term} and \emph{short-term} learning objectives, while $M_I^{(t)}$ and $M_E^{(t)}$ represent \emph{implicit} and \emph{explicit} motivational components. Each set contains zero or more components, where each component $c$ is a structured object populated by an LLM, defined by the following attributes:
\begin{equation}
\label{eq:component_structure}
c = \{\text{description, metric, evidence, confidence, status}\},
\end{equation}
where \texttt{status} can be either \texttt{[NOT\_ALIGNED]} or \texttt{[ALIGNED]}. All components in the initial state $s_0$ are initialized as \texttt{[NOT\_ALIGNED]}. We provide an example of this structured state in Table~\ref{tab:learner_state_example}.

The state then evolves dynamically throughout the learning path. After each learning action $k_t$, the learner's interactions $I_t$ are processed by an evaluator LLM $F_\text{eval}$, acting as a state transition function that updates the prior state $s_{t}$ to a new state $s_{t+1}$:
\begin{equation}
\label{eq:state_transition}
s_{t+1} = F_{\text{eval}}(s_{t}, I_t),
\end{equation}
where $F_{\text{eval}}$ (1) updates \texttt{status} of existing components $c \in s_{t}$ that meet their predefined metrics, and (2) introduces new components emerging from interaction evidence. This iterative process yields a trajectory of evolving states $(s_0, s_1, \dots, s_T)$ that provides a dynamic foundation for policy learning. The specific prompts used for this state extraction and evaluation are detailed in Appendix~\ref{appendix:system_prompt}.

\subsubsection{Automated Abstract Reward Function}
\label{sec:reward_function}

To guide the reinforcement learning, we define a scalar abstract reward function $\mathcal{R}(s_t, k_{t+1})$ that quantifies progress toward predefined pedagogical goals.

Let $C(s)$ be the set of all components present in a state $s$. For each component $c \in C(s)$, we define two functions:

$\bullet$ An \textbf{indicator function}, $\phi: \mathcal{S} \times C \to \{0, 1\}$, which returns $1$ if $c$ has status \texttt{[ALIGNED]} in state $s$, and $0$ otherwise.

$\bullet$ A \textbf{confidence function}, $\text{conf}: \mathcal{S} \times C \to [0, 1]$, which returns the evaluator's confidence score in component correctness.

The reward for taking action $k_{t+1}$ is computed as the weighted sum of newly aligned components in $s_t$:
\begin{equation}
\label{eq:reward_function}
\mathcal{R}(s_t, k_{t+1}) =
\sum_{c \in C(s_{t+1})}
w_c \cdot \text{conf}(s_{t+1}, c)
\cdot \big[\phi(s_{t+1}, c) - \phi(s_t, c)\big],
\end{equation}
where $s_{t+1} \sim T(\cdot | s_t, k_{t+1})$ is the resulting state transition, and $w_c \ge 0$ is the empirical weight assigned to component $c$.
The key term $\phi(s_{t+1}, c) - \phi(s_t, c)$ ensures that only transitions that change a component’s status from \texttt{[NOT\_ALIGNED]}to \texttt{[ALIGNED]} contribute positively, directly rewarding goal advancement. This formulation directly incentivizes the policy to select actions that advance the learner state with respect to their objectives and motivations.

\begin{table*}[t]
    \centering
    \caption{An example of the structured Learner State Model. The model is organized into four key dimensions, each of which can contain multiple trackable items representing the learner's goals and motivations.}
    \label{tab:learner_state_example}
    \begin{tabularx}{\textwidth}{@{}ccX@{}}
        \toprule
        \textbf{Dimension} & \textbf{Item} & \textbf{Details} \\
        \midrule
        \textbf{Long-Term} & Objective 1 & \textbf{Description:} Understand the concepts of BP and PDP. \\
        \textbf{Objective} & ($O_L^{(1)}$) & \textbf{Metric:} concept\_understanding\_score (\textbf{Threshold:} $\ge$0.90) \\
        & & \textbf{Measurement:} Assesses in-session deep understanding of BP and PDP concepts. \\
        & & \textbf{Evidence:} [Turn 36: "What are BP and PDP?"] \\
        & & \textbf{Confidence:} 0.65 \quad \textbf{Status:} [NOT\_ALIGNED] \\
        \midrule
        \textbf{Short-Term} & Objective 1 & \textbf{Description:} Grasp basic AI concepts during the session. \\
        \textbf{Objective} & ($O_S^{(1)}$) & \textbf{Metric:} immediate\_concept\_recall (\textbf{Threshold:} $\ge$0.80) \\
        & & \textbf{Measurement:} Assesses immediate recall of AI concepts (e.g., self-supervision, MCTS). \\
        & & \textbf{Evidence:} [Turn 43: "What is Monte Carlo Tree Search?"] \\
        & & \textbf{Confidence:} 0.75 \quad \textbf{Status:} [NOT\_ALIGNED] \\
        \midrule
        \textbf{Implicit} & Motivation 1 & \textbf{Description:} Desire for autonomous control over the course content. \\
        \textbf{Motivation} & ($M_E^{(1)}$) & \textbf{Metric:} control\_preference\_score (\textbf{Threshold:} $\ge$0.70) \\
        & & \textbf{Measurement:} Based on the frequency of requests to skip content and interventions in the lesson's pace. \\
        & & \textbf{Evidence:} [Turn 20: "Skip this page, teacher."] \\
        & & \textbf{Confidence:} 0.80 \quad \textbf{Status:} [NOT\_ALIGNED] \\
        \midrule
        \textbf{Explicit} & Motivation 1 & \textbf{Description:} Interest in the impact of technological advancements. \\
        \textbf{Motivation} & ($M_I^{(1)}$) & \textbf{Metric:} technology\_impact\_attention (\textbf{Threshold:} $\ge$0.80) \\
        & & \textbf{Measurement:} Based on the frequency of questions about technological progress and its societal impact. \\
        & & \textbf{Evidence:} [Turn 25: "The development of AI is changing... impact..."] \\
        & & \textbf{Confidence:} 0.68 \quad \textbf{Status:} [NOT\_ALIGNED] \\
        \bottomrule
    \end{tabularx}
\end{table*}

\subsubsection{Policy Training}
\label{sec:policy_learning}

To obtain a policy capable of robust, long-horizon planning, we employ a two-stage training paradigm integrating Supervised Fine-Tuning (SFT) and Goal-driven Relative Policy Optimization (GRPO).

\paragraph{\textbf{Stage 1: Supervised Fine-Tuning (SFT) for Initialization}}
This stage initializes the policy to produce pedagogically valid actions based on expert demonstrations.

We construct a high-quality dataset $\mathcal{D} = \{(s, k^*)\}$ from a real-world learning system~\cite{yu2024moocmaicreshapingonline}, where each record contains a \emph{structured learner state} and the \emph{expert-preferred action}. The curation process involves: (1) collecting learner interactions from $300$ learning sessions with $14,584$ interaction turns (each session includes one or more learning action); (2) leveraging GPT-4o~\cite{openai2024gpt4ocard} to process and generate a structured learner state $s$ for each learning session; and (3) employing three human experts to identify the optimal next action, $k^*$. This resulted in a dataset with real learner data and expert demonstrations. Detailed statistics are provided in Table~\ref{tab:dataset-statistics}.

Based on this dataset, we fine-tune the initial policy parameters $\theta$ by minimizing the negative log-likelihood of the expert actions:
\begin{equation}
\label{eq:sft_loss}
\mathcal{L}_{\text{SFT}}(\theta)
= - \mathbb{E}{(s, k^*) \sim \mathcal{D}}
\big[\log \pi_\theta(k^* | s)\big].
\end{equation}
This behavioral cloning yields an initial policy to make locally optimal decisions that align with expert judgment.

\paragraph{\textbf{Stage 2: Group Relative Policy Optimization (GRPO) for Long-Horizon Planning}}
While SFT ensures short-term correctness, it lacks long-horizon planning. We further refine the policy using GRPO, a variance-reduced gradient method that stabilizes learning under heterogeneous reward distributions.

Given a batch of trajectories $\tau=\{(s_t, k_{t+1},\mathcal{R_t})\}$ sampled from the current policy $\pi_\theta$, the objective is:
\begin{equation}
\label{eq:grpo_objective}
J_{\text{GRPO}}(\theta)
= \mathbb{E}_{\tau \sim \pi_\theta}
\left[
\frac{\pi_\theta(k_{t+1} | s_t)}{\pi_{\theta_{\text{old}}}(k_{t+1} | s_t)}
\cdot \hat{A}_t^{\text{grp}}
\right],
\end{equation}
where $\hat{A}^{\text{grp}}_t$ is the group-relative advantage:
\begin{equation}
\hat{A}_t^{\text{grp}} = \frac{A_t - \mu_g}{\sigma_g + \epsilon},
\quad A_t = R_t + \gamma V(s{t+1}) - V(s_t).
\end{equation}
Here, $\mu_g$ and $\sigma_g$ are the mean and standard deviation of advantages within a sampled group of trajectories $g$.

By combining SFT initialization and GRPO refinement, the resulting policy achieves both local pedagogical coherence and global goal alignment, effectively producing adaptive and interpretable learning paths across diverse learner profiles.

\begin{table}[t]
    \centering
    \small
    \caption{Overview of the learner interaction dataset, including number of  learning sessions, interaction turns, and the composition of the learner states.}
    \label{tab:dataset-statistics}
    \begin{tabular}{lcccccc} 
        \toprule
        & & & \multicolumn{4}{c}{\textbf{\# Learner States}} \\ 
        \cmidrule(lr){4-7}
         & \textbf{\# Session} & \textbf{\# Interaction} & \textbf{\# $O_L$} & \textbf{$O_S$} & \textbf{$M_I$} & \textbf{$M_E$} \\
        \midrule
        Train & $250$ & $11,328$ & $273$ & $325$ & $275$ & $288$ \\
        Test & $50$ & $3,256$ & $53$ & $76$ & $63$ & $62$  \\
        \midrule
        Total & $300$ & $14,584$ & $326$ & $401$ & $338$ & $350$ \\
        \bottomrule
    \end{tabular}
\end{table}

\subsection{LLM-driven Educational Architecture}

To validate and operationalize the trained policy model, we design an LLM-driven educational architecture that integrates structured learner profiling, learning path planning and adaptive delivery. This architecture provides real-time, context-sensitive learning experience that evolves with learner's behavioral and cognitive state.

\subsubsection{Pre-planning Architecture}
\label{sec:perception_module}

Before policy-driven planning, the system performs real-time learner profiling and candidate retrieval to construct an adaptive action space.

\begin{figure}[ht]
\centering
\includegraphics[width=0.9\columnwidth]{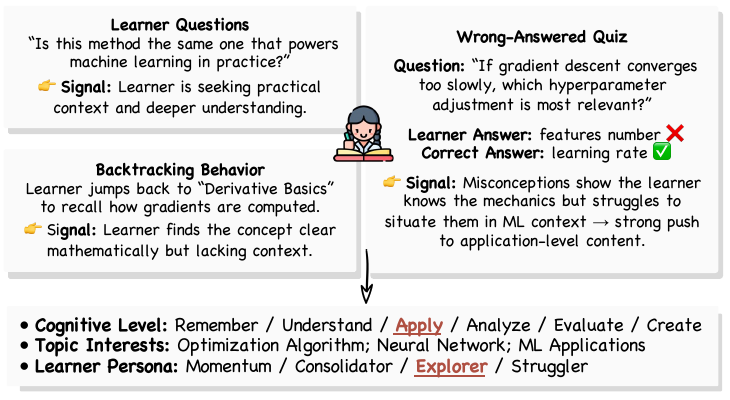}
  \caption{Example of \system's pre-planning architecture that generates a structured learner profile.}
  \label{fig:perception_module}
\end{figure}

\textbf{Learner Profiling.} The profiling pipeline consists of three sequential stages that process diverse learning traces to infer behavioral, cognitive, and motivational characteristics (Figure~\ref{fig:perception_module}).

\textbf{$\bullet$ Stage 1: Behavioral Pattern Analysis.}
Raw interaction logs, including navigation traces, page dwell times, review sequences, and quiz outcomes—are transformed into structured behavioral indicators via LLM-based analysis. For instance, long dwell times suggest sustained engagement, while rapid transitions may imply distraction. Similarly, repeated revisits to the same material indicate consolidation strategies, and clusters of incorrect quiz responses reveal misconceptions. This provide interpretable metrics of \textbf{engagement}, \textbf{review behavior}, and \textbf{conceptual understanding}, forming the behavioral foundation of the learner profile.

\textbf{$\bullet$ Stage 2: Semantic and Intent Analysis.}
At this stage, learner discussions and written inputs are analyzed to extract semantic and affective signals. At the micro level, each message is annotated with (1) \textbf{cognitive type} (e.g., remembering, applying, analyzing) following Bloom’s taxonomy~\cite{bloom1956}, (2) \textbf{affective state} (e.g., confused, confident, motivated), and (3) \textbf{communicative intent} (e.g., questioning, reflecting, disagreeing). At the macro level, discourse aggregation yields higher-order attributes, such as \textbf{thematic coherence}, \textbf{cognitive progression}, and \textbf{interaction dynamics}, providing a structured interpretation of unstructured dialogue data.

\textbf{$\bullet$ Stage 3: Synthesis and Profile Formulation.}
Signals from previous stages are then synthesized into a concise learner profile. The system infers the learner’s current \textbf{cognitive level} (e.g., Understanding, Applying, Analyzing) and core \textbf{topic interests} using linguistic cues, emotional markers, and review behaviors. Based on this synthesis, learners are dynamically classified into four prototypical personas: (1) \textbf{Momentum Learner} (rapid progress), (2) \textbf{Consolidator} (focused on review and mastery), (3) \textbf{Explorer} (branching into new areas), or (4) \textbf{Struggler} (requiring remedial guidance). The resulting profile $p_t$ at time $t$ encapsulates the learner’s cognitive state, motivational orientation, and interest distribution:
\begin{equation}
\small
p_t = f_{\text{profile}}(I_t) = \{\text{cognition}, \text{engagement}, \text{interest}, \text{persona}\},
\end{equation}
where $I_t$ denotes the interaction data for session $t$. This profile serves as input to the subsequent planning. Implementation details and system prompts are provided in Appendix~\ref{appendix:system_prompt}.

\textbf{Retrieval.}
Once the profile is constructed, the system retrieves a manageable set of pedagogically relevant actions. We employ a hybrid retrieval mechanism that integrates \textbf{lexical relevance} and \textbf{semantic similarity}, with the relevance score computed as:
\begin{equation}
\text{score}(k) = \alpha \cdot \text{BM25}(p_t, k) + (1-\alpha) \cdot \text{sim}_{\text{dense}}(p_t, k),
\end{equation}
where $\text{BM25}$ is the keyword-matching score~\cite{robertson2009probabilistic}, and $\text{sim}_{\text{dense}}$ is the cosine similarity from dense embeddings. After excluding actions already taken, the top-k items (empirically $k = 10$) form the candidate action set $\mathcal{K}_c \subseteq \mathcal{K}$.

\textbf{Learning Path Planning.} 
Given the current learner state $s_t$ and candidate actions $\mathcal{K}_c$, the policy $\pi_\theta$ evaluates each candidate and selects the optimal next action:
\begin{equation}
k_{t+1}^* = \arg\max_{k \in \mathcal{K}c} \mathbb{E}_{\pi_\theta}[\mathcal{R}(s_t, k) + \gamma V(s_{t+1})],
\end{equation}
where $V(s_{t+1})$ is the estimated value of the next state and $\mathcal{R}(s_t, k)$ is the expected pedagogical reward.
Unlike retrieval-only baselines, this policy explicitly optimizes for long-term cumulative reward, ensuring coherent and pedagogically grounded path planning.

\subsubsection{Post-Planning Architecture}
\label{sec:action_module}

To ensure a seamless and contextually consistent learning experience, we design a post-planning delivery architecture that transforms discrete policy outputs into coherent instructional narratives. 

\textbf{Adaptive Narrative Generation.}
Instead of displaying the selected action, the system generates a narrative bridge connecting the new content to the learner’s ongoing context.
Given the learner profile $p_t$, learning history $H_t = \{k_1, \dots, k_t\}$, and selected action $k_{t+1}^*$, the delivery model $\mathcal{D}(\cdot)$ generates personalized instructions:
\begin{equation}
o_t = \mathcal{D}(p_t, H_t, k_{t+1}^*),
\end{equation}
where the output $o_t$ adapts tone, emphasis, and phrasing to align with the learner’s cognitive and motivational state.
For instance, a \emph{Momentum Learner} might receive succinct, challenge-oriented phrasing, whereas a \emph{Struggler} might receive supportive scaffolding referencing past misconceptions. In summary, our post-planning architecture yields pedagogically coherent transitions that maintain engagement while ensuring conceptual continuity.

\textbf{$\bullet$ Expandable Architecture.}
Our post-planning architecture is designed as a modular and extensible component, allowing integration of additional generative services for richer instructional experiences including \textbf{Retrieval-Augmented Generation (RAG)} for in-depth explanations or \textbf{Multimodal Generation}. As a proof of concept, we implement a dynamic slide generator that synthesizes new slides from selected actions while preserving stylistic and formatting consistency. This maintains visual and cognitive coherence across materials and reduces the cognitive load associated with inconsistent presentation formats. 
We leave the comprehensive exploration of such generative extensions to future work.

\definecolor{RedBase}{HTML}{C72324}
\definecolor{BlueBase}{HTML}{2978B5}

\colorlet{QwenLight}{RedBase!15}   
\colorlet{LlamaLight}{BlueBase!18} 

\newcolumntype{G}{>{\columncolor{gray!20}}c}
\begin{table*}[t]
    \centering
    \small
    \caption{Performance evaluated by average pedagogical alignment rate (\%), reward computation ($\mathcal{R(\cdot)}$) within the defined Learner State Model. \system consistently achieves the highest alignment rate and reward values.}
    \begin{tabular}{l|ccccG|ccccG|ccccG}
    \toprule
    \textbf{Model} & \multicolumn{5}{c|}{\textbf{Alignment Rate (\%)}} & \multicolumn{5}{c|}{\textbf{Reward Computation ($\mathcal{R(\cdot)}$)}} & \multicolumn{5}{c}{\textbf{\# Total Component}} \\
     & $O_L$ & $O_S$ & $M_I$ & $M_E$ & Avg  & $\mathcal{R}(O_L)$ & $\mathcal{R}(O_S)$ & $\mathcal{R}(M_I)$ & $\mathcal{R}(M_E)$ & Total & $\#O_L$ & $\#O_S$ & $\#M_I$ & $\#M_E$ & Total \\
    \midrule
    GPT-4o                 & $$53.45$$ & $$40.23$$ & $$69.23$$ & $$51.47$$ & $$52.52$$ & $31.0$ & $34.8$ & $45.0$ & $35.0$ & $145.8$ & $58$ & $87$ & $65$ & $68$ & $278$ \\
    4o-mini                & $$37.93$$ & $$23.46$$ & $$49.23$$ & $$35.29$$ & $$35.66$$ & $22.0$ & $19.0$ & $32.0$ & $24.0$ & $97.0$  & $58$ & $81$ & $65$ & $68$ & $272$ \\
    \cellcolor{QwenLight}Qwen3
                           & $$37.70$$ & $$31.82$$ & $$49.28$$ & $$41.89$$ & $$39.73$$ & $23.0$ & $28.0$ & $34.0$ & $31.0$ & $116.0$ & $61$ & $88$ & $69$ & $74$ & $292$ \\
    \cellcolor{LlamaLight}Llama3.1
                           & $$36.21$$ & $$32.56$$ & $$56.72$$ & $$49.25$$ & $$43.17$$ & $21.0$ & $28.0$ & $38.0$ & $33.0$ & $120.0$ & $58$ & $86$ & $67$ & $67$ & $278$ \\
    \midrule
    GPT-4o$_{Infer}$       & $$61.29$$ & $$40.91$$ & $$74.63$$ & $$62.50$$ & $$58.48$$ & $38.0$ & $36.0$ & $50.0$ & $45.0$ & $169.0$ & $62$ & $88$ & $67$ & $72$ & $289$ \\
    4o-mini$_{Infer}$      & $$50.00$$ & $$39.08$$ & $$70.15$$ & $$53.73$$ & $$52.33$$ & $29.0$ & $34.0$ & $47.0$ & $36.0$ & $146.0$ & $58$ & $87$ & $67$ & $67$ & $279$ \\
    \cellcolor{QwenLight}Qwen3$_{Infer}$
                           & $$56.67$$ & $$36.47$$ & $$63.24$$ & $$57.97$$ & $$52.48$$ & $34.0$ & $31.0$ & $43.0$ & $40.0$ & $148.0$ & $60$ & $85$ & $68$ & $69$ & $282$ \\
    \cellcolor{LlamaLight}Llama3.1$_{Infer}$
                           & $$49.18$$ & $$34.09$$ & $$63.77$$ & $$64.00$$ & $$51.88$$ & $30.0$ & $30.0$ & $44.0$ & $48.0$ & $152.0$ & $61$ & $88$ & $69$ & $75$ & $292$ \\
    \midrule
    \cellcolor{QwenLight}Qwen3$_{SFT}$
                           & \underline{$$65.00$$} & \underline{$$55.81$$} & $$71.01$$ & $$60.87$$ & $$62.68$$ & \textbf{$$39.0$$} & \underline{$48.0$} & $49.0$ & $42.0$ & $178.0$ & $60$ & $86$ & $69$ & $69$ & $284$ \\
    \cellcolor{LlamaLight}Llama3.1$_{SFT}$
                           & $$60.66$$ & $$44.83$$ & \underline{$$76.12$$} & $$62.86$$ & $$60.00$$ & $37.0$ & $39.0$ & \underline{$51.0$} & \underline{$44.0$} & $171.0$ & $61$ & $87$ & $67$ & $70$ & $171$ \\
    \midrule
    \cellcolor{QwenLight}Pxplore$_{Qwen3}$
                           & \textbf{$$66.10$$} & \textbf{$$58.33$$} & \underline{$$76.12$$} & \underline{$$63.24$$} & \textbf{$$65.47$$} & \textbf{$$39.0$$} & \textbf{$$49.0$$} & \underline{$51.0$} & $43.0$ & \textbf{$$182.0$$} & $59$ & $84$ & $67$ & $68$ & $278$ \\
    \cellcolor{LlamaLight}Pxplore$_{Llama3.1}$
                           & $$63.33$$ & $$47.67$$ & \textbf{$$80.30$$} & \textbf{$$67.14$$} & \underline{$$63.48$$} & \underline{$38.0$} & $41.0$ & \textbf{$$53.0$$} & \textbf{$$47.0$$} & \underline{$179.0$} & $60$ & $86$ & $66$ & $70$ & $282$ \\
    \bottomrule
    \end{tabular}
    \label{tab:method_results}
\end{table*} 

\section{Experiment}

In this section, we conduct extensive experiments to evaluate the effectiveness of \system’s training paradigm and the educational architecture from macro to micro perspectives.

\subsection{Training Paradigm Evaluation}
We first conduct an end-to-end validation experiment using the curated dataset described in Section~\ref{sec:policy_learning}, to evaluate the effectiveness of our reward function and training paradigm. 

\textbf{Experimental Setups.}
Building on our curated dataset $\mathcal{D}$ (Table~\ref{tab:dataset-statistics}), we compare our trained policy model ($Pxplore_\text{Base}$) with three groups of baselines: (1) \textbf{Prompt-based LLMs}, serving as zero-shot baselines to test out-of-the-box planning capabilities of modern LLMs, including \emph{GPT-4o}~\cite{openai2024gpt4ocard}, \emph{GPT-4o-mini}~\cite{openai2024gpt4ocard}, \emph{Qwen3-8B}~\cite{qwen3technicalreport}, and \emph{Llama3.1-8B-Instruct}~\cite{llama318binstructtechnicalreport}. (2) \textbf{Inference-Guided LLMs}, which adopt the same prompt-based approach but include a predefined Learner State Model as additional context for decision-making. (3) \textbf{SFT} models, including \emph{Qwen3-8B}, and \emph{Llama3.1-8B-Instruct} fine-tuned on our annotated state-action dataset. 

\textbf{Results.}
Each model is evaluated on its ability to select the optimal action from the candidate set given a learner state. We then measure the resulting alignment with predefined pedagogical objectives. Table~\ref{tab:method_results} shows \system's consistent and substantial improvements in overall pedagogical alignment rate. \textbf{Prompt-based} baselines, despite their strong general reasoning abilities, often fail to make decisions that align with multi-dimensional educational goals. Adding \textbf{inference-time guidance} through structured learner states yields noticeable improvements, yet remains limited by models' inherent short-sightedness. Within our training paradigm, \textbf{SFT}-tuned models produce a clear boost by enabling locally optimal decisions, while  \textbf{GRPO} refinement achieves the best performance across all metrics. Notably, our GRPO-trained \emph{Qwen3-8B} reaches an overall alignment rate of $65.47\%$, outperforming the much larger proprietary model, \emph{GPT-4o} even with inference-time guidance. These results validate the effectiveness of our reinforcement-based paradigm in optimizing abstract educational rewards and promoting long-horizon, goal-aligned learning path planning.

\subsection{Modular Evaluation}
Following the validation of the overall training paradigm, we conduct a fine-grained modular evaluation of each functional module within the \system framework, including the modules of (1) learner profiling, (2) learning path planning, and (3) adaptive delivery. 

\begin{figure}[ht]
  \centering
  \includegraphics[width=\columnwidth]{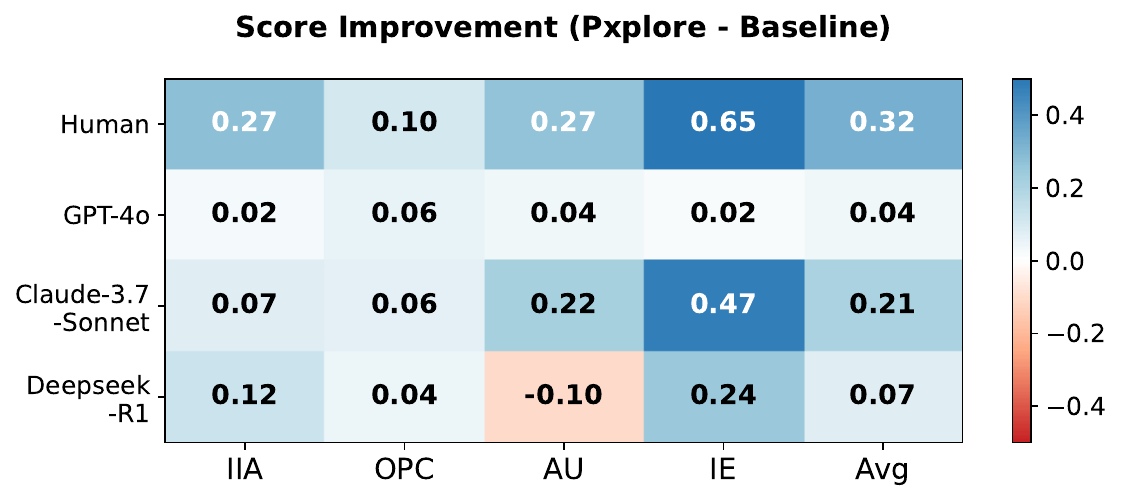}
  \caption{Heatmap showing the improvement in evaluation scores on a 5-point Likert scale across pedagogical dimensions, calculated as \(\text{Score}_{\system} - \text{Score}_{\text{Baseline}}\).}
  \label{fig:perception_result}
\end{figure}

\begin{figure*}[t]
  \centering
  \includegraphics[width=0.95\linewidth]{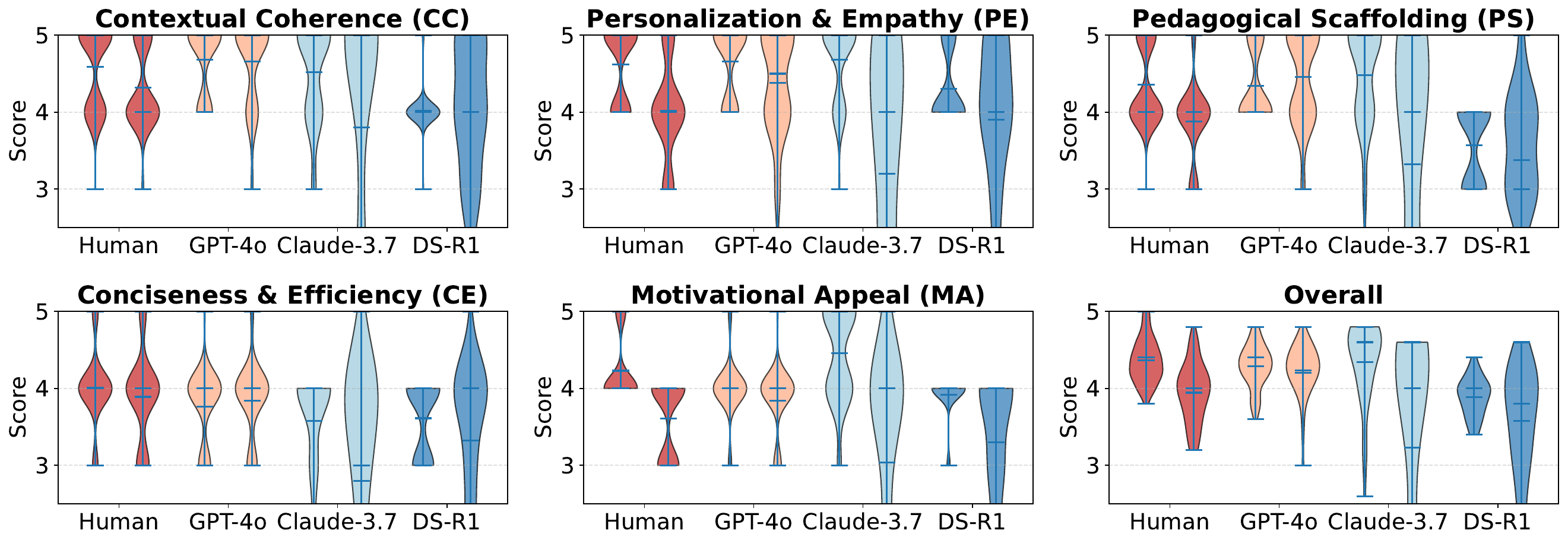}
  \caption{ Evaluation scores (1–5) across pedagogical dimensions. For each violin pair (Human, \emph{GPT-4o}, \emph{Claude-3.7-Sonnet}, and \emph{Deepseek-R1}), the left represents \system’s adaptive delivery and the right represents the original content.}
  \label{fig:result_action}
\end{figure*}

\subsubsection{Profiling Module}
To evaluate the quality of the profiling module in \system pre-planning archiecture, we adopt a hybrid approach that integrates human judgment and LLM-as-a-judge~\cite{gu2024survey}.

\textbf{Experimental Setup.}  Using the curated dataset introduced in the previous section, we generate a structured learner profile for each learning session following the three-stage framework in Section~\ref{sec:perception_module}. These profiles are compared against GPT-4o baseline, which recieves the same raw session logs but only a single instruction to generate a profile following the same output schema. Each profile is independently rated by two human annotators and three LLM-judges (\emph{Claude-3.7-Sonnet}~\cite{anthropic2025claude37}, \emph{GPT-4o}, and \emph{Deepseek-R1}~\cite{deepseekai2025}) on a 5-point Likert scale, all blind to source. Evaluations are provided across four pedagogical dimensions: 
(1) \textbf{\emph{Interest Identification Accuracy (IIA)}}: The correctness and specificity of inferred topic interests in the generated profile.
(2) \textbf{\emph{The overall Profile Completeness (OPC)}}: Coverage of core attributes, including engagement, cognition, affect, and interests.
(3) \textbf{\emph{Actionability and Utility (AU)}}: The practical usefulness of the generated profile for guiding downstream planning.
(4) \textbf{\emph{Interpretability and Explainability (IE)}}: Clarity of reasoning and strength of evidence linking learner behaviors to inferred states.

\textbf{Results.}
To ensure reliability, we aggregate results from both human annotators and LLM-judges (Figure~\ref{fig:perception_result}). Human annotators consistently rate \system higher across all dimensions, with strong inter-annotator agreement (Pearson correlation $r>0.8$). Notably, \system achieved the largest improvement in \textbf{IE} ($+17.7\%$), followed by \textbf{IIA} ($+6.26\%$) and \textbf{AU} ($+6.63\%$), resulting in an overall increase of $+7.88\%$. Both human and LLM-judges highlight \system's \emph{ability to construct explicit "evidence chain"}, for example, linking a learner's review behavior directly to a detected misconception. This structured interpretability is essential for trustworthy learner profiling and aligns with established educational principles. 
Among LLM-judges, we observe different scoring patterns. \emph{Claude-3.7-Sonnet} is the most conservative, while \emph{Deepseek-R1} shows highest alignment with human judgment.
Despite the differences, all evaluators confirm a consistent performance gap, validating the accuracy, completeness, and interpretability of \system's profiling framework.

\begin{figure}[t]
  \centering
  \includegraphics[width=0.95\columnwidth]{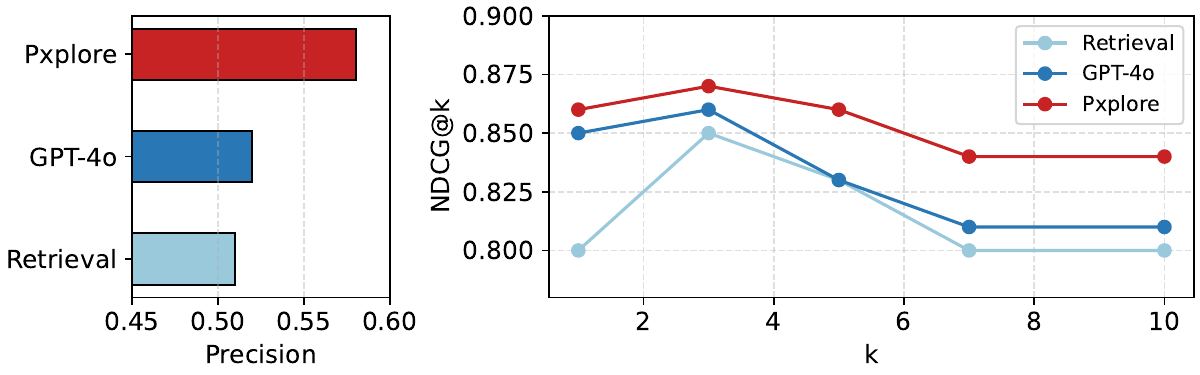}
  \caption{Results on the alignment precision of top-ranked action with the expert-annotated "best" choice, as well as the overall ranking quality across the candidate set.}
  \label{fig:decision_ndcg}
\end{figure}

\subsubsection{Planning Module}
We next evaluate the planning module, focusing on its alignment with expert pedagogical judgment.

\textbf{Data Construction.}
We construct a fine-grained knowledge corpus by segmenting $148$ courses from a real-world learning platform~\cite{yu2024systematic} into atomic actions (coherent learning units derived from slides and transcripts). Each action is enriched with metadata, including a summary, keyword tags, and a Bloom's Taxonomy~\cite{bloom1956} level (e.g., Understanding, Analyzing), generated by GPT-4o and verified by human experts for pedagogical soundness. For each learning session, we link the generated profile with the corresponding learning history to form a dataset containing the profile, candidate set, and expert-ranked actions. Three pedagogical experts independently label each candidate as \emph{best}, \emph{acceptable}, or \emph{not suitable} to establish a ground truth for the evaluation. 

\textbf{Experimental Setup.}
We compare the trained \system policy model with two baselines: (1) a \textbf{retrieval-only} mechanism (Section~\ref{sec:perception_module}) and (2) \textbf{GPT-4o} prompted under identical conditions. Performance is evaluated using two complementary metrics:

\noindent(1) \textbf{\emph{Precision@1 (P@1)}}: The agreement between the top-ranked action and the expert-annotated \emph{best} choice).

\noindent(2) \textbf{\emph{Normalized Discounted Cumulative Gain (NDCG@k)}}: The overall ranking quality across the candidate set.

\textbf{Results.}
The results are shown in Figure~\ref{fig:decision_ndcg}, \system achieves the strongest alignment with expert judgment. In terms of \textbf{P@1}, \system's top-ranked selection matches the expert-annotated "best" choice most frequently, outperforming both GPT-4o and the retrieval-only baseline. Furthermore, the \textbf{NDCG@k} results demonstrate that \system consistently yields higher ranking quality across all evaluated list depths ($k=1, 3, 5, 7, 10$), indicating that \system not only identifies the single best option but also effectively prioritizes all pedagogically appropriate alternatives.

\subsubsection{Delivery Module}
We evaluate the delivery module which transforms discrete actions into contextually coherent instructions. 

\textbf{Experimental Setup.} We use the dataset from prior evaluations and compare \system's adapted content against the original materials. Two human evaluators and three LLM-judges assess each version on a 5-point Likert scale across five pedagogical dimensions:
(1) \textbf{\emph{Contextual Coherence (CC)}}: How the content is logically structured and flows smoothly from the prior context.
(2) \textbf{\emph{Personalization \& Empathy (PE)}}: Appropriateness and alignment with the learner’s current goals, knowledge state, and needs.
(3) \textbf{\emph{Pedagogical Scaffolding (PS)}}: Effectiveness of the content in providing structured guidance and support for learning.
(4) \textbf{\emph{Conciseness \& Efficiency (CE)}}: Clarity, brevity, and effectiveness of the language in conveying key ideas.
(5) \textbf{\emph{Motivational Appeal (MA)}}: Capacity to inspire and sustain learner motivation and engagement.

\textbf{Results.}
Figure~\ref{fig:result_action} shows that \system consistently outperforms the original content. The largest gains appear in \textbf{MA} ($+17.2\%$), \textbf{PE} ($+14.9\%$) and \textbf{PS} ($+12.4\%$), leading to an overall improvement of $+10.7\%$. We attribute this to \system's ability to \emph{generate "narrative bridges", the contextual transition that address misconceptions or sustain engagement}. 
Additionally, the alignment between LLM-judges and human evaluators further reinforces the robustness of our results. While the improvement in \textbf{CE} is marginal ($+3.1\%$), and GPT-4o even rates the original as more concise, this reflects a design trade-off: \system deliberately prioritizes empathy and motivation over minimal brevity to achieve richer pedagogical impact. Importantly, we ensure that total word counts of the adapted content remain comparable through targeted condensation of redundant segments, ensuring efficient yet engaging delivery.

\subsection{Real-World Experiment}

To validate the practical effectiveness of \system framework in authentic learning contexts, we integrate it into a deployed online learning platform and conducted a controlled user study.

\textbf{Experimental Setup}. We recruit 22 undergraduate students ($N=22$) from diverse academic backgrounds and randomly assigned them into two groups: (1) the \emph{Experimental Group} ($n=11$), using the full \system framework, and (2) the \emph{Control Group} ($n=11$), using a retrieval-based content-matching system representative of existing adaptive learning platforms. All participants begins with a validated introductory lesson: \emph{Towards Artificial General Intelligence (TAGI)}, selected for its high-quality instructional design and broad accessibility. After completing the initial lesson, participants in experimental group proceed through the full planning pipeline powered by \system's policy, while those in control group follow the next lesson recommended by retrieval-based path planning.

\textbf{Procedure.} Each learning session lasts approximately one hour, designed to simulate a realistic learning context: 
(1) \textbf{Pre-Test (10 mins)}: Participants complete a diagnostic test of 10 multiple-choice questions related to the initial lesson. 
(2) \textbf{Initial Learning Session (30 mins)}: Participant engage with one complete lesson on the system at their own pace. 
(3) \textbf{Advanced Learning Session (15 mins)}: After completing the initial lesson, participants proceed through the next lesson determined by the assigned planning method. 
(4) \textbf{Post-Test and Survey (10 mins)}: Participants complete a follow-up assessment and a perception survey. 

\textbf{Metrics.} We evaluate both learning outcomes and learner experience. \textbf{Learning outcomes} are measured by pre- and post-test score improvement across multiple-choice questions and short-answer questions, while \textbf{learner experience} is assessed using a 5-point Likert-scale across following dimensions: 
(1) \textbf{\emph{Helpfulness (Help.)}}: Perceived usefulness of learning path; 
(2) \textbf{\emph{Diversity (Div.)}}: Variety in topics and formats; 
(3) \textbf{\emph{Relevance (Rel)}}: Alignment with prior learning content;
(4) \textbf{\emph{Personalization (Pers.}}: Adaptation to individual preferences; 
(5) \textbf{\emph{Clarity (Clar.)}}: Learning path coherence and logical flow; 
(6) \textbf{\emph{Motivation (Mot.)}}: Stimulation of interest and engagement; 
(7) \textbf{\emph{Understanding (Und.)}}: Depth of comprehension gained; 
(8) \textbf{\emph{Satisfaction (Sat.)}}: The overall satisfaction score.

\begin{figure}[t]
  \centering
  \includegraphics[width=0.95\columnwidth]{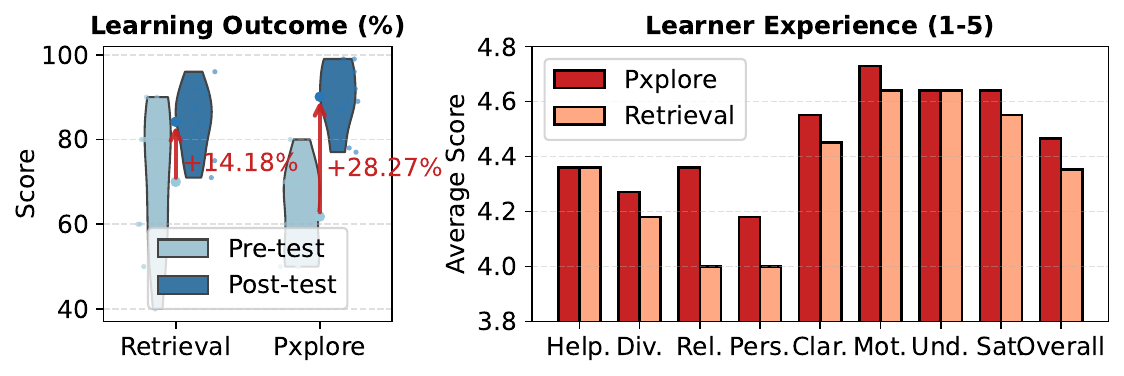}
  \caption{Comparison of \system and retrieval-based baseline in terms of learning outcomes and learner experience.}
  \label{fig:result_user_study}
\end{figure}

\textbf{Results.}
As shown in Figure~\ref{fig:result_user_study}, participants using \system achieve significantly greater knowledge gains than those using the retrieval-based baseline. Specifically, \system improves mean test scores from $61.81\%$ (pre-test) to $90.09\%$ (post-test), resulting in a knowledge gain of $+28.28\%$, compared with a smaller increase observed in the control group. Although both groups ultimately achieve comparable post-test performance, the experimental group exhibits a steeper improvement that suggests more effective conceptual progression.

Additionally, Figure~\ref{fig:result_user_study} further shows that \system's advantage in subjective learner experience. Participants rate \system notably higher in \textbf{Relevance} ($+0.36$) and \textbf{Personalization} ($+0.18$), confirming its ability to construct learning paths that better reflect prior context and individual preferences. The most pronounced performance gaps appear in \textbf{Motivation} ($4.73$) and \textbf{Satisfaction} ($4.64$), indicating that the goal-driven, coherent paths produced by \system foster more engaging and enjoyable learning experiences. Interestingly, both systems achieve similar ratings in \textbf{Helpfulness} and \textbf{Understanding}, likely due to the foundational nature of the initial instructional content and the comparable effort invested across conditions. This aligns with the post-test parity, implying that \system’s advantage lies in both marginal performance and the quality and engagement of the learning experience itself.


\section{Conclusion}

In this paper, we present \system, a reinforcement learning paradigm that leverages an automated pedagogical reward function to transform complex learner states into computable optimization signals. Through comprehensive experiments, we demonstrate that \system consistently outperforms baselines in guiding learners toward their long-term developmental goals. By deploying \system's trained policy model within a real-world learning platform, which integrates a real-time learner profiling before planning, and an adaptive content delivery after planning. This further validates \system's capability to construct coherent and goal-aligned learning paths.


\begin{acks}
This work is supported by National Natural Science Foundation of China (No.62277033) and National Engineering Laboratory for Cyberlearning and Intelligent Technology. We also thank the Deng Feng Fund of Tsinghua University for supporting this research.
\end{acks}

\bibliographystyle{ACM-Reference-Format}
\bibliography{sample-base}

\appendix

\section{System Prompts}
\label{appendix:system_prompt}

In this section, we present the system prompts used throughout our \system framework. These prompts guide the LLMs in extracting learner states, planning personalized learning paths, and delivering adaptive content. Full prompt details are publicly available in the repository referenced in this paper.

\subsection{Learner State Extraction and Evaluation}
Table~\ref{tab:learner_state_extraction} provides system prompts used to extract and evaluate learner states by guiding the model to infer learners’ long-term and short-term objectives, explicit motivations, and implicit motivations., as described in Section~\ref{sec:learner_state_model}. 

\begin{table}[h]
\centering
\small
\caption{Prompts for learner state extraction and evaluation.}
\label{tab:learner_state_extraction}
\begin{tabular}{p{\linewidth}}
\toprule
\multicolumn{1}{c}{\textbf{Learner State Extraction}} \\
\midrule
\textbf{Role}: You are a senior expert in educational psychology and computational linguistics, specializing in building structured student profiles by extracting assessable cognitive, emotional, and motivational features from student interactions. This is based on cutting-edge learning science theories (such as Bloom's Taxonomy, Webb's Depth of Knowledge Model, and Pekrun's Achievement Emotion Theory) and multi-turn dialogue analysis. \\
\textbf{Task Description:} Based on the dialogue history of student \{student\_name\} with the teacher/peer intelligent agents, provide an analysis from the following five dimensions: state\_description, long\_term\_objective, short\_term\_objective, implicit\_motivation, explicit\_motivation. \\
\textbf{Layout}: \{Input and Output Constraints\} \\
\midrule
\multicolumn{1}{c}{\textbf{Learner State Evaluation}} \\
\midrule
\textbf{Role}: You are a senior expert in educational psychology and computational linguistics, specializing in building and updating structured student profiles by extracting assessable cognitive, emotional, and motivational features from student interactions in online classrooms, based on cutting-edge learning science theories and multi-turn dialogue analysis. \\
\textbf{Task Description:} \\
$\bullet$ Re-evaluate all existing items in the initial\_state using combined evidence from both the initial state and the next\_lesson\_content. This includes updating the evidence list, re-calculating the metric, and setting the is\_aligned flag to true or false based on the specified threshold. \\
$\bullet$ Identify and add new objectives or motivations if supported by new evidence from the next\_lesson\_content, following the original micro-template. \\
$\bullet$ Update the state\_description to summarize key changes compared to the initial\_state, citing significant new behavioral evidence from next\_lesson\_content. \\
$\bullet$ Update the confidence score (0.00-1.00) for each item based on the quality and directness of the available evidence. \{Detailed Confidence Update Rules\} \\
\textbf{Layout}: \{Input and Output Constraints\} \\
\bottomrule
\end{tabular}
\end{table} 

\subsection{Module Prompts}
\begin{figure}[ht]
  \centering
  \includegraphics[width=0.9\columnwidth]{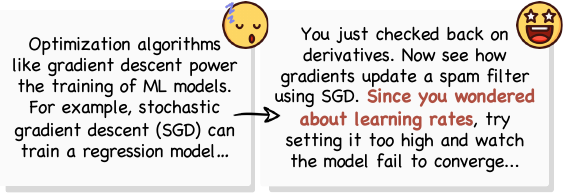}
  \caption{Example of \system's post-planning architecture that enables adaptive delivery of selected action, connecting the new content to the learner's on-going context.}
  \label{fig:sample-adaptation}
\end{figure}

We further include the system prompts employed in each major module of our framework: learner profiling (Table~\ref{tab:prompt_profiling}), learning path planning (Table~\ref{tab:prompt_planning}), and adaptive delivery (Table~\ref{tab:prompt_delivery}). Figure~\ref{fig:sample-adaptation} shows an example of content generated by the adaptive delivery module.

\begin{table}[h]
\centering
\small
\caption{System prompts for two-stage adaptive delivery.}
\label{tab:prompt_delivery}
\begin{tabular}{p{\linewidth}}
\toprule
\multicolumn{1}{c}{\textbf{Stage 1: Suggestion Generation}} \\
\midrule
\textbf{Role}: You are a professional assistant for optimizing academic content. \\
\textbf{Task}: Based on the provided inputs, generate a set of modification suggestions for the next lecture script to better align with the personal needs. \\
\textbf{Requirements} \\
$\bullet$ Briefly summarize the core ideas and logical flow of the current content. \\
$\bullet$ Propose personalized, actionable modification directions (e.g., adding case studies, reinforcing a specific knowledge point) by integrating the recommendation rationale and the recommended outline. \\
$\bullet$ Modifications must be targeted optimizations of the original content, not replacements. The original sequence of topics must be maintained. \\
\textbf{Layout}: \{Input and Output Constraints\} \\
\midrule
\multicolumn{1}{c}{\textbf{Stage 2: Script Generation}} \\
\midrule
\textbf{Role}:
You are an expert educational content creator and scriptwriter. \\
\textbf{Task}: Rewrite the original lecture script by seamlessly integrating the provided modification suggestions. \\
\textbf{Requirements}: \{Detailed Requirements\} \\
\textbf{Layout}: \{Input and Output Constraints\} \\
\bottomrule
\end{tabular}
\end{table}

\begin{table}[h]
\centering
\small
\caption{System prompts for three-stage profiling.}
\label{tab:prompt_profiling}
\begin{tabular}{p{\linewidth}}
\toprule
\multicolumn{1}{c}{\textbf{Stage 1: Behavioral Analysis}} \\
\midrule
\textbf{Role}: You are a professional agent combining the expertise of a learning analyst and an educational data mining researcher. \\
\textbf{Task}: Your goal is to receive a raw JSON object representing a student’s learning session, perform a deep analysis on three specific areas (page\_interactions, review\_loops, and quizzes), and output the analysis.\\
\textbf{Step 1: Analyze Page Interactions}: \{Detailed Anomaly and Rules\} \\
\textbf{Step 2: Analyze Review Loops}: \{Detailed Anomaly and Rules\} \\
\textbf{Step 3: Analyze Quizzes}: \{Detailed Anomaly and Rules\} \\
\textbf{Step 4: Construct the Final Analysis}: \{Detailed Anomaly and Rules\} \\
\textbf{Layout}: \{Input and Output Constraints\} \\
\midrule
\multicolumn{1}{c}{\textbf{Stage 2: Semantic Analysis}} \\
\midrule
\textbf{Role}: You are a educational psychology analyst and computational linguist, skilled at transforming student discussion logs into structured, insightful diagnostic data. Your analysis aims to uncover the cognitive processes, affective states, and communicative intents behind student utterances. \\
\textbf{Task}: Process a complete, structured student learning session log. Your core task is to perform a two-tiered analysis on the discussion\_threads array: a micro-level analysis of individual student messages and a macro-level analysis of each discussion thread as a whole. \\
\textbf{Step 0: Global Context Awareness}: \{Detailed Anomaly and Rules\} \\
\textbf{Step 1: Message Analysis (Micro-level)}: \{Detailed Anomaly and Rules\} \\
\textbf{Step 2: Thread Analysis (Macro-level)}: \{Detailed Anomaly and Rules\} \\
\textbf{Layout}: \{Input and Output Constraints\} \\
\midrule
\multicolumn{1}{c}{\textbf{Stage 3: Profile Formulation}} \\
\midrule
\textbf{Role}: You are a top-tier expert in personalized learning content recommendation, proficient in learning science and data analysis. \\
\textbf{Task}: Your task is to generate 2-4 high-value learning content recommendations for a student based on a pre-processed log of learning episodes \\
\textbf{Global Analysis \& State Determination}: \{Detailed Anomaly and Rules\} \\
\textbf{Core Element Extraction}: \{Detailed Anomaly and Rules\} \\
\textbf{Core Recommendation Strategy}: \{Detailed Anomaly and Rules\} \\
\textbf{Layout}: \{Input and Output Constraints\} \\
\bottomrule
\end{tabular}
\end{table}

\begin{table}[h]
\centering
\small
\caption{System prompt for learning path planning.}
\label{tab:prompt_planning}
\begin{tabular}{p{\linewidth}}
\toprule
\multicolumn{1}{c}{\textbf{Learning Path Planning}} \\
\midrule
\textbf{Role}: You are an intelligent educational assistant. Your task is to plan a personalized learning path for a student by selecting the next instructional content snippet based on their current learning situation. \\
\textbf{Task Description}: Combine the student's recommendation\_strategy with a comprehensive analysis of each candidate snippet to select the one most suitable for the current student and instructional context: \\
$\bullet$ recommendation\_strategy: A personalized recommendation basis derived from interaction history, behavioral analysis, and other information. \\
$\bullet$ candidates: Several candidate content snippets retrieved by the system, containing fields such as title, keywords, Bloom's level, and summary. \\
$\bullet$ Relevance with the current content to ensure logical coherence. \\
$\bullet$ Appropriateness for the ability level (e.g., Bloom's cognitive level). \\
\textbf{Layout}: \{Input and Output Constraints\} \\
\bottomrule
\end{tabular}
\end{table}

\section{Hyperparameters}
\label{appendix:hyperparameter}

In this section, we provide a comprehensive summary of the hyperparameter settings for our training paradigm, as well as the configurations (Table~\ref{tab:training-config-sft}) used in the retrieval process (Table~\ref{tab:retrieval-config}).

\begin{table}[h]
\centering
\caption{Training setups and hyperparameters used in our two-stage training paradigm.}
\label{tab:training-config-sft}
\begin{tabular}{|c|c|}
\toprule
\textbf{Parameter$_{SFT}$} & \textbf{Value} \\
\midrule
Learning Rate & $5.0 \times 10^{-6}$ \\
Number of Epochs & $$10$$ \\
Max Sequence Length & $$2048$$ \\
Warmup Ratio & $$0.1$$ \\
Temperature & $$0.5$$ \\
\midrule
\textbf{Parameter$_{GRPO}$} & \textbf{Value} \\
\midrule
Learning Rate & $1 \times 10^{-6}$ \\
Batch Size & $$4$$ \\
Number of Epochs & $$10$$ \\
Max Sequence Length & $$8192$$ \\
Max Completion Length & $$1024$$ \\
Warmup Ratio & $$0.05$$ \\
Temperature & $$0.5$$ \\
\bottomrule
\end{tabular}
\end{table}

\begin{figure}[h!]
  \centering
  \includegraphics[width=0.9\columnwidth]{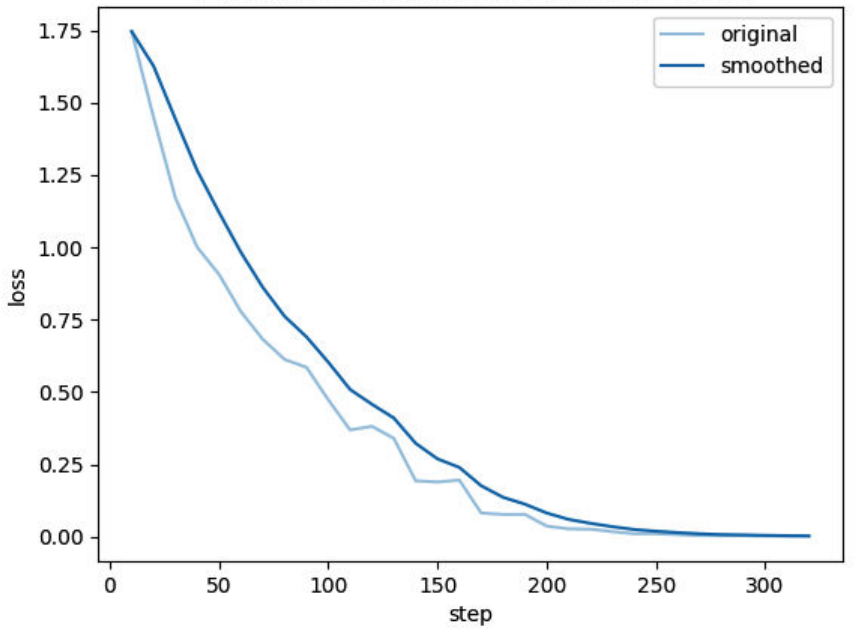}
  \caption{Training loss curve of \textbf{Qwen3-8B}.}
  \label{fig:training_loss_qwen3}
\end{figure}

\begin{figure}[h!]
  \centering
  \includegraphics[width=0.9\columnwidth]{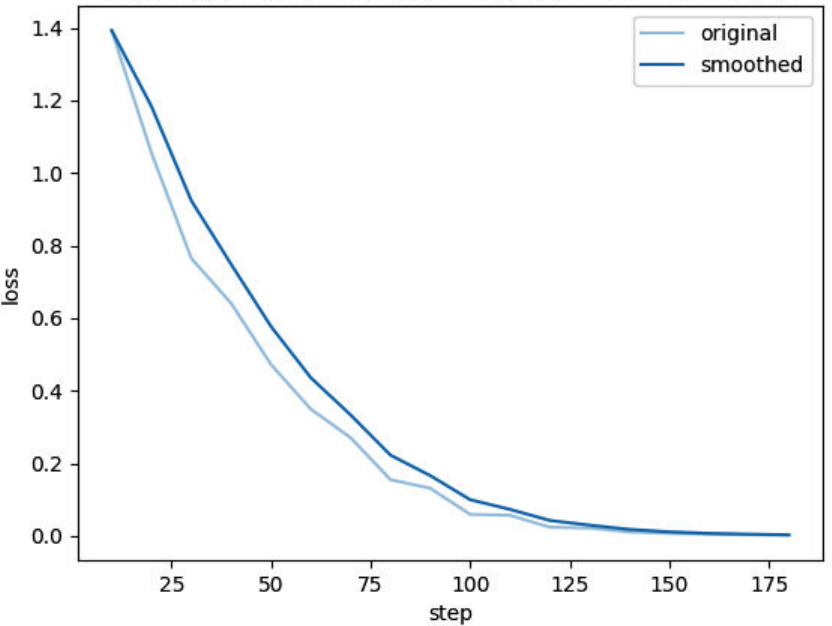}
  \caption{Training loss curve of \textbf{Llama-3.1-Instruct}.}
  \label{fig:training_loss_llama}
\end{figure}

\begin{table*}[t!]
\centering
\small
\caption{Pre- and Post-Test questions to measure knowledge gain after the learning process.}
\begin{tabular}{cp{0.95\linewidth}}
\toprule
\textbf{\#} & \textbf{Pre-Test Multi-Choice Questions} \\ 
\midrule
1 & In artificial intelligence, what does the term \textit{modality} refer to? \\ 
2 & When teaching an AI model the concept of ``cat,'' which approach would provide the most comprehensive and robust understanding? \\ 
3 & In deep neural networks, what is \textit{convolution} operation? \\ 
4 & Comparing the tasks of ``drawing'' a cat vs. ``recognizing'' a cat, which is generally considered more challenging? \\ 
5 & In audio processing, what is the purpose of converting a continuous waveform into a spectrogram of frequency and intensity over time? \\ 
6 & When an AI model can process text, images, sound, and other modalities together, what is the greatest advantage of this ability? \\ 
7 & In multimodal models (e.g., CLIP), what does the core objective \textit{cross-modal alignment} mean? \\ 
8 & If a LLM (e.g., ChatGPT) can accurately generate images from text descriptions, what capability does this most likely imply? \\
9 & When teaching a machine the concept of ``apple,'' which approach provides the most single-dimensional information? \\
10 & In AI, the Transformer architecture first became famous for achieving breakthrough progress in which domain? \\ 
\midrule
\textbf{\#} & \textbf{Post-Test Multi-Choice Questions} \\ 
\midrule
1 & Which of the following belongs to the field of ``multimodal'' research? \\ 
2 & Compared with traditional CNN models, what is one main advantage of the ViT model? \\ 
3 & In the course, the method of ``starting from random noise and progressively denoising to generate an image'' was compared to what? \\ 
4 & What is the main difference between OpenAI's Whisper model and the Suno model? \\ 
5 & For models such as CLIP and ImageBind, what is the purpose of the core objective ``multimodal alignment''? \\ 
6 & In ResNet, what core problem does the introduction of ``residual connections'' aim to solve? \\ 
7 & In the course, the ``pooling'' layer of CNNs was introduced as simulating what function or role in the biological visual system? \\ 
8 & What is the primary function of the pooling layer in a convolutional neural network (CNN)? \\ 
9 & In ``deep diffusion models'' (e.g., Stable Diffusion), what is the core step followed during image generation? \\ 
10 & In training multimodal LLMs (e.g., those handling text, images, and audio), what is one major advantage in terms of data usage? \\ 
\midrule
\textbf{\#} & \textbf{Post-Test Short Answer Questions} \\ 
\midrule
1 &	In the course, we mentioned text-to-image or text-to-video models (e.g., Sora). What is one fundamental capability they all rely on? Why is this capability so crucial? Please provide a reasonable explanation. \\
2 & After completing this course, what do you think is the greatest value or potential of multimodal large models (e.g., those that can understand both images and text) compared to single-task models like image recognition systems or chatbots? Please illustrate with an example.\\
\bottomrule
\end{tabular}
\label{tab:test_questions}
\end{table*}

\begin{table}[h]
\centering
\caption{Embedding model and retrieval configurations.}
\label{tab:retrieval-config}
\begin{tabular}{cc}
\toprule
\textbf{Component} & \textbf{Configuration} \\
\midrule
Embedding Model & \texttt{baai/bge-large-zh-v1.5}. \\
Similarity Metric & Cosine similarity \\
Weight Coefficient ($\alpha$) & $0.2$ (empirically tuned) \\
Top-$k$ Retrieval & $k=10$ \\
\bottomrule
\end{tabular}
\end{table}

\begin{table}[ht]
\centering
\small
\caption{Survey questions to measure learner experience on a 5-point Likert-scale.}
\begin{tabular}{cp{0.9\linewidth}}
\toprule
\textbf{\#} & \textbf{Learner Experience Questionnaire} \\ 
\midrule
1 & I think the content recommended is very helpful for my learning. \\ 
2 & The system’s recommended content is sufficiently diverse in topics and formats, so it does not feel monotonous. \\ 
3 & The system’s recommended content is well connected to what I have previously learned. \\ 
4 & The system can provide personalized recommendations based on my learning preferences. \\ 
5 & I think the learning path guided by the system is clear and logical. \\ 
6 & This system has sparked my interest in further exploring the field. \\ 
7 & I feel I have gained a deeper understanding of the learning content after using this system. \\ 
8 & I would be willing to recommend this system to my classmates. \\ 
\bottomrule
\end{tabular}
\label{tab:experience_questionnaire}
\end{table}

\section{Real-World Experiment}
\label{appendix:user_study}

We include supplementary materials for the real-world experiment to facilitate the replication of our study and enable further analysis. To ensure a thorough evaluation of the learning outcomes, all participants completed pre- and post-tests, followed by a detailed perception questionnaire aimed at assessing their overall learning experience and engagement with the intervention.

Table~\ref{tab:test_questions} presents the complete set of test questions used in the experiment. The test consists of 10 multiple-choice questions (MCQs) and 2 short-answer questions (SAQs), covering cognitive levels from \textit{Remembering} to \textit{Applying} according to Bloom’s taxonomy. The overall test score is computed as:
$$
\text{Score} = 0.8 \times \text{Score}_{\text{MCQ}} + 0.2 \times \text{Score}_{\text{SAQ}}
$$
This weighting balances factual recall and open-ended reasoning to better reflect holistic learning outcomes.

Table~\ref{tab:experience_questionnaire} lists the learner experience questionnaire used for subjective evaluation. Participants rated each item on a 5-point Likert scale ranging from “Strongly Disagree” to “Strongly Agree.” The questionnaire covers eight pedagogical dimensions, helpfulness, diversity, relevance, personalization, clarity, motivation, understanding, and satisfaction, providing a comprehensive assessment of perceived adaptivity and engagement. 

To ensure transparency and facilitate future research, all anonymized survey responses and test results are made publicly available in our repository: ~\url{https://github.com/Pxplore/pxplore-algo}.

\end{document}